\documentclass{article}

\PassOptionsToPackage{numbers, compress}{natbib}
\usepackage[preprint]{neurips_2025}

\usepackage{CJK}
\usepackage{listings}

\usepackage[utf8]{inputenc} 
\usepackage[T1]{fontenc}    
\usepackage{hyperref}       
\usepackage{url}            
\usepackage{booktabs}       
\usepackage{amsfonts}       
\usepackage{nicefrac}       
\usepackage{microtype}      
\usepackage{xcolor}         
\usepackage{tabularx}
\usepackage{fontawesome}
\usepackage{pifont}

\usepackage{enumitem}

\usepackage{graphicx} 
\usepackage{subcaption} 

\definecolor{pythonblue}{RGB}{84, 184, 255}

\usepackage[most]{tcolorbox}
\usepackage{lipsum}
\usepackage{geometry}

\usepackage{upquote}
\usepackage{listings}
\usepackage{ulem}

\usepackage{tcolorbox}
\newtcolorbox{problembox}{colback=blue!5!white, colframe=blue!75!black, fonttitle=\bfseries, title=Sample Problem}
\newtcolorbox{codebox}{colback=pythonblue!5!white,
colframe=pythonblue!75!black, fonttitle=\bfseries, title=Canonical Solution}

\usepackage{multirow}

\makeatletter
\DeclareRobustCommand\onedot{\futurelet\@let@token\@onedot}
\def\@onedot{\ifx\@let@token.\else.\null\fi\xspace}

\makeatother

\title{OIBench: Benchmarking Strong Reasoning Models with Olympiad in Informatics}

\author{%
  Yaoming Zhu$^{1}$,\, Junxin Wang$^{2}$,\, Yiyang Li$^{3}$,\, Lin Qiu$^{3}$,\, \textbf{ZongYu Wang}$^{3}$,\,\and  \textbf{Jun Xu}$^{3}$, \textbf{Xuezhi Cao}$^{3}$,\, \textbf{Yuhuai Wei}$^{3}$,\, \textbf{Mingshi Wang}$^{3}$,\, \textbf{Xunliang Cai}$^{3}$,\, \and \textbf{Rong Ma}$^{4}$\\
  \\
  $^{1}$AGI-Eval, $^{2}$Beijing Normal University,  $^{3}$Meituan,  $^{4}$Shanghai Jiao Tong Univerisity\\
}

\begin{document}
\begin{CJK}{UTF8}{gbsn}

\maketitle

\begin{abstract}
  As models become increasingly sophisticated, conventional algorithm benchmarks are increasingly saturated, underscoring the need for more challenging benchmarks to guide future improvements in algorithmic reasoning. This paper introduces OIBench, a high-quality, private, and challenging olympiad-level informatics dataset comprising 250 carefully curated original problems. We detail the construction methodology of the benchmark, ensuring a comprehensive assessment across various programming paradigms and complexities, and we demonstrate its contamination-resistant properties via experiments. We propose Time/Space Completion Curves for finer-grained efficiency analysis and enable direct human-model comparisons through high-level participant evaluations. Our experiments reveal that while open-source models lag behind closed-source counterparts, current SOTA models already outperform most human participants in both correctness and efficiency, while still being suboptimal compared to the canonical solutions. By releasing OIBench as a fully open-source resource\footnote{\url{https://huggingface.co/datasets/AGI-Eval/OIBench}}, we hope this benchmark will contribute to advancing code reasoning capabilities for future LLMs.

\end{abstract}

\section{Introduction}

Large language models have demonstrated remarkable capabilities in algorithmic reasoning and complex problem-solving, and significantly empowered human productivity in these tasks, assisting programmers in designing algorithms, optimizing computational efficiency, and solving high-level programming challenges. 
As models become increasingly sophisticated, conventional algorithm benchmarks like HumanEval~\citep{DBLP:journals/corr/abs-2107-03374}  and MBPP~\citep{austin2021program} are saturated, with most state-of-the-art LLMs achieving solve rates exceeding 90\%. The saturation underscores the need for more challenging benchmarks to guide future improvements in algorithmic reasoning~\citep{DBLP:journals/corr/abs-2410-05229,DBLP:journals/corr/abs-2410-23123}.

Recent benchmarks proposed more difficult coding datasets by collecting data from open-source platforms, such as Codeforces~\citep{DBLP:journals/corr/abs-2203-07814,DBLP:journals/corr/abs-2501-01257}, USACO~\citep{DBLP:journals/corr/abs-2404-10952}, and LeetCode~\citep{DBLP:journals/corr/abs-2403-07974,DBLP:conf/nips/0005QSCZ24}. However, their test data faced significant exposure risks given that modern LLMs utilize large-scale Internet data as their pre-training corpus. In particular, many large models will extensively crawl data from code competitions to improve performance~\citep{DBLP:conf/ease/CoignionQR24}.

The emergence of recent advanced test-time scaling models, such as DeepSeek-R1~\citep{guo2025deepseek} and OpenAI-O1~\citep{jaech2024openai}, has further elevated the reasoning capabilities of LLMs, especially on complex problems. Their technical reports often compare the performances with human, as CodeElo's~\citep{DBLP:journals/corr/abs-2501-01257} online rating strategy.
However, these evaluation schemes require models to participate in online competitions over months. Moreover, human participants' varying numbers and skill levels in different sessions introduce data noise and reproducibility challenges.
Although EffiBench introduces execution time and memory usage metrics, these remain coarse-grained average scores that obscure critical performance variations across problem types.

To address the limitations, we present OIBench, a high-quality, private, and challenging informatics olympiad-level dataset comprising 250 carefully curated original problems. The OIBench is bilingual; each problem is presented in Chinese and English with rigorous verification confirming their absence from public repositories before release.
We also conduct experiments to show that supervised training cannot solve such in-distribution challenges, preventing leakage risks from the dataset open-source to future dataset updates. To enable finer-grained efficiency analysis, we propose Time/Space Completion Curves, which visualize algorithmic optimality across varying time/space limits, surpassing the granularity of prior normalized averages.
In addition, we invite participants of high-level programming competitions to solve a subset of OIBench, enabling direct comparisons between LLMs and human performance. To foster reproducibility, OIBench is fully open-source, releasing not only standard components (problems, test cases, difficulty tags) but also canonical solutions, baseline model responses, and detailed reproduction costs.

We list the OIBench's contributions as follows: 
\begin{enumerate}[left=0pt]
\item OIBench features highly private and challenging data, collected exclusively from coaches of OI competitions. We verify through search engine queries that none of these problems are publicly accessible before release. Furthermore, our experiments demonstrate OIBench's resilience against data leakage: conventional models fail to acquire substantial problem-solving capabilities on our benchmark even when fine-tuned with in-distribution data. 
\item We measure LLM's ability to solve Olympiad-level coding problems via OIBench, where strong reasoning models significantly outperform conventional LLMs. We develop additional pseudo-code evaluation based on the original problems to systematically assess different models' coding capabilities. Our paper presents an in-depth analysis of these evaluations.
\item We also compare the SOTA models' performance with ACM-level human participants on a subset of OIBench, and conduct a comprehensive analysis. We find that current strong reasoning models outperform average ACM-level contestants in both time and space complexity when solving algorithmic problems, even reaching optimal levels.
\item OIBench is fully open-source, releasing not only standard components (problems, labels, difficulty levels, test cases) but also canonical solutions. Considering the accessibility and reproducibility challenges of closed-source models, we further provide all baseline model responses and detailed reproduction costs.

\end{enumerate}

\section{Related Work}

\textbf{Reasoning Benchmarks}. Since recent studies~\citep{wei2022chain} introduced the concept of Chain-of-Thought~(CoT), the research community has regarded it as a plausible way towards strong artificial intelligence~\citep{huang2023towards,xu2025towards}. Based on CoT, researchers developed various benchmarks on measuring the LLM's reasoning abilities, including mathematics (GSM8K~\citep{DBLP:journals/corr/abs-2110-14168}, MATH~\citep{DBLP:conf/nips/HendrycksBKABTS21}), commonsense(HotpotQA~\citep{DBLP:conf/emnlp/Yang0ZBCSM18} ) , logics~\citep{DBLP:conf/acl/ParmarPVN0MMB24} , coding   and comprehensive ones(CriticBench~\citep{DBLP:conf/acl/LinGLLLY24}).

Recent studies debated whether existing benchmarks properly assessed reasoning capabilities. While some attributed improvements to memorization~\citep{DBLP:journals/corr/abs-2410-05229}, others attributed insufficient challenging problems in benchmarks~\citep{DBLP:journals/corr/abs-2410-23123}. The community developed new benchmarks to address these gaps: Sys2Bench (multi-task reasoning)~\citep{DBLP:journals/corr/abs-2502-12521}, CriticBench (critique/correction)~\citep{DBLP:conf/acl/LinGLLLY24}, LogicVista (visual logic)~\citep{DBLP:journals/corr/abs-2407-04973}, LiveBench (contamination-free evaluation)~\citep{DBLP:journals/corr/abs-2406-09170}, LogicGame (rule-based planning)~\citep{DBLP:journals/corr/abs-2408-15778}, and QRData (statistical reasoning)~\citep{DBLP:conf/acl/LiuW0LCF24}. These collectively provided systematic evaluation frameworks while revealing LLMs' reasoning strengths and weaknesses.

\textbf{Coding Benchmarks}, such as HumanEval~\citep{DBLP:journals/corr/abs-2107-03374} and CoderEval~\citep{DBLP:conf/icse/YuSRZZMLLWX24}, offered the standard for evaluating the code capabilities of models via programming problems. More complex benchmarks, such as CodeElo~\citep{DBLP:journals/corr/abs-2501-01257} and USACO~\citep{DBLP:journals/corr/abs-2404-10952}, introduced more challenging, competition-level problems. EffiBench~\citep{DBLP:conf/nips/0005QSCZ24} focused on evaluating the efficiency of the generated code, while the RACE benchmark further extended the evaluation dimensions by considering four key aspects\citep{DBLP:journals/corr/abs-2407-11470,DBLP:journals/corr/abs-2301-09043}. CRUXEval~\citep{DBLP:conf/icml/GuRLSS024} and EquiBench~\citep{DBLP:journals/corr/abs-2502-12466} assess LLM performance in code understanding tasks. 
PseudoEval~\citep{DBLP:journals/corr/abs-2502-19149} evaluates models' coding ability via pseudocode. Existing static benchmarks were susceptible to data contamination, compromising the reliability and generalizability of evaluation results.
DynaCode~\citep{DBLP:journals/corr/abs-2503-06643} effectively mitigated data contamination by generating new code samples~\citep{DBLP:journals/corr/abs-2503-04149}. Additionally, LiveCodeBench~\citep{DBLP:journals/corr/abs-2403-07974} provided a comprehensive, low-contamination code evaluation platform, enhancing the timeliness of benchmarks.




\section{OIBench Construction}

The olympiad-level informatics problems are originally collected from ACM-ICPC team coaches. These coaches, on average, have 20 years of experience in coaching university ACM teams and high school Olympiad Informatics (OI) teams in China. Their expertise in creating high-difficulty algorithmic problems is instrumental in our data collection process.

To ensure the quality of these challenging algorithmic problems in our study, we ask the coaches either to select problems from their private repositories or to compose new problems that meet our stringent criteria as follows:
\begin{enumerate}[left=0pt]
    \item \textbf{Originality and Confidentiality}:  To prevent data leakage, each problem must be original and unpublished, whether online or in print-media. Importantly, minor adaptations of existing problems, such as changing numerical values or rephrasing the problem statement without altering the underlying logic, are not considered original. 
    \item \textbf{Appropriate Difficulty Level}: To ensure that the problems provided by the coaches are difficult enough to effectively assess frontier models' reasoning capabilities, We require coaches to assign difficulty labels to problems based on the Codeforces difficulty rating. In addition, each problem must be labeled with its corresponding competition level. 
    
    
    \item \textbf{Robust Test Cases and Verified Canonical Solutions}: Effective test cases are essential for accurately evaluating the participants' solutions. Usually, by including test scenarios against large input sizes and resource-intensive operations, test cases can reveal efficiencies and potential bottlenecks in time and space utilization for a solution program. Besides, high-quality test cases shall also examine the program's resilience against corner cases. 
    Hence, we require the coaches to craft test cases that meet our standard listed in the Appendix. Additionally, coaches are required to provide canonical solutions in \textbf{C++} to confirm that all problems are solvable and test cases are correct. We also verify the correctness of the canonical programs by executing all test cases. 
  
\end{enumerate}

We employ six Informatics Olympiad participants to review the quality of the questions after the collection process, ensuring that each question is \textit{solvable}, \textit{clearly described}, includes \textit{appropriate sample input/output}, and that the formulas/tables in the problem statement \textit{adhere to \TeX/Markdown syntax}. 

\begin{figure}
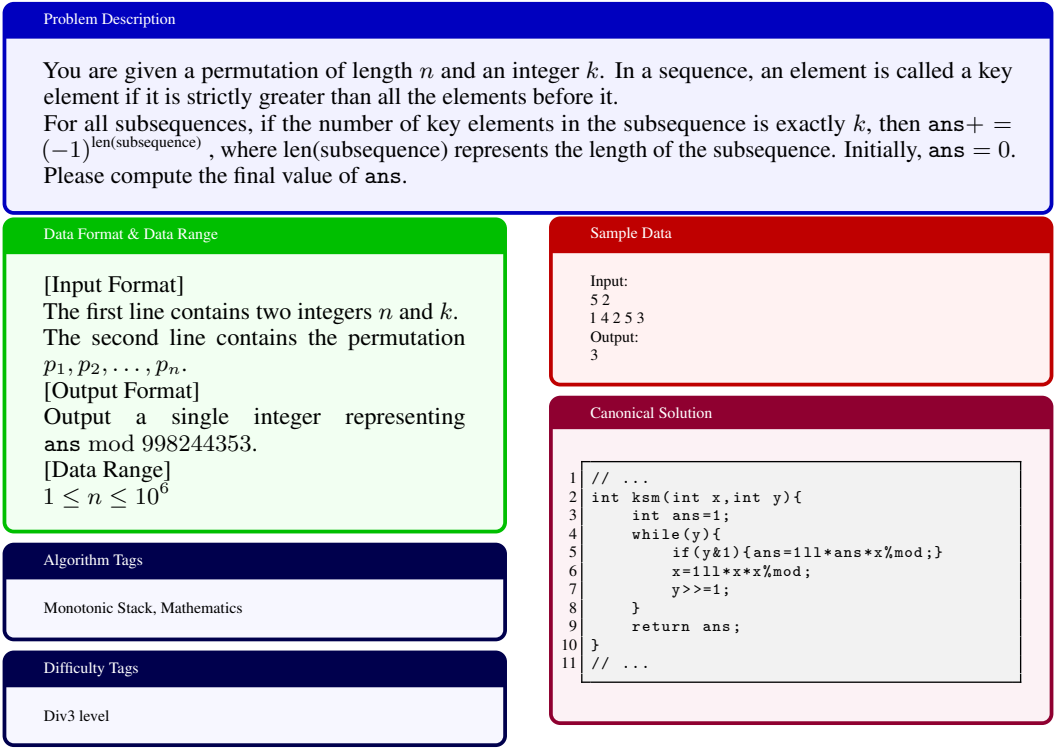

\tiny
\begin{minipage}[t]{1\textwidth}
\tcbset{colback=blue!5!white, colframe=blue!75!black}
\begin{tcolorbox}[title=Problem Description]
\small

You are given a permutation of length $n$ and an integer $k$.
In a sequence, an element is called a key element if it is strictly greater than all the elements before it.

For all subsequences, if the number of key elements in the subsequence is exactly $k$, then $\mathtt{ans} += (-1)^{\text{len(subsequence)}}$
, where $\text{len(subsequence)}$ represents the length of the subsequence.
Initially, $\mathtt{ans}=0$. 

Please compute the final value of $\mathtt{ans}$.

\end{tcolorbox}
\end{minipage}

\begin{minipage}[t]{0.48\textwidth}

\tcbset{colback=green!5!white, colframe=green!75!black}
\begin{tcolorbox}[title=Data Format \& Data Range ]
\small
[Input Format]

The first line contains two integers $n$ and $k$.

The second line contains the permutation $p_1, p_2, \ldots, p_n$.

[Output Format]

Output a single integer representing  $\mathtt{ans} \bmod 998244353$.

[Data Range]

$1 \leq n \leq 10^6$
\end{tcolorbox}

\tcbset{colback=blue!3!white, colframe=blue!30!black}
\begin{tcolorbox}[title=Algorithm Tags]
Monotonic Stack, Mathematics
\end{tcolorbox}

\tcbset{colback=blue!3!white, colframe=blue!30!black}
\begin{tcolorbox}[title=Difficulty Tags]
Div3 level
\end{tcolorbox}

\end{minipage}
\hfill
\begin{minipage}[t]{0.48\textwidth}
\tcbset{colback=red!5!white, colframe=red!75!black}
\vspace{-20em}
\begin{tcolorbox}[title=Sample Data]
Input:

5 2

1 4 2 5 3

Output:

3

\end{tcolorbox}

\tcbset{colback=purple!5!white, colframe=purple!75!black}

\begin{tcolorbox}[title=Canonical Solution]
\tiny
\begin{lstlisting}
// ...
int ksm(int x,int y){
    int ans=1;
    while(y){
        if(y&1){ans=1ll*ans*x%mod;}
        x=1ll*x*x%mod;
        y>>=1;
    }
    return ans;
}
// ...
\end{lstlisting}
\end{tcolorbox}

\end{minipage}
\caption{Example problem.}
\label{fig/sample}
\vspace{-0.5cm}
\end{figure}

After verifying the quality of the Chinese problem statements, we recruit professional translators with related work experience to translate the problems into English. During this process, we involve ACM coaches and informatics competition participants in providing an Informatics Olympiad terminology glossary, which we also open-source for the research community.
We include the background profiles, work duration, and wage information of all dataset annotation employees in the Appendix~\ref{sec:Recruitment}.
We give an example problem in Fig.~\ref{fig/sample}.


\begin{table}[h]
\tiny
\centering
\setlength{\tabcolsep}{2.5pt}
\begin{tabular}{c|ccccccccc}
\hline
                                                                              & Diffculty        & \# problems & \begin{tabular}[c]{@{}c@{}}avg. word\\ per prob\end{tabular} & \begin{tabular}[c]{@{}c@{}}avg.size of\\ test per prob\end{tabular} & \begin{tabular}[c]{@{}c@{}}avg. lines of \\ canonical solution\end{tabular} &  non-symbolic & \begin{tabular}[c]{@{}c@{}}time/space\\ metrics\end{tabular} & language & source     \\ \hline
HumanEval\citep{DBLP:journals/corr/abs-2107-03374}     & $\star$             & 164         & 67.7                                                         & 65.4k                                                               & 8.7                                                                         &  \ding{55}           &  \ding{55}                                                           & en       & Private    \\
CodeContests\citep{DBLP:journals/corr/abs-2203-07814}  & $\star$$\star$         & 13.3k       & 344.5                                                        & 91.7k                                                               & 52.5                                                                        &  \ding{51}          &  \ding{55}                                                           & en       & Codeforces \\
USACO\citep{DBLP:journals/corr/abs-2404-10952}         & $\star$$\star$$\star$     & 307         & 452.9                                                        & 59.8k                                                               & 28.7                                                                        &  \ding{51}          &  \ding{55}                                                           & en       & USACO      \\
CodeElo\citep{DBLP:journals/corr/abs-2501-01257}       & $\star$$\star$         & 387         & 147.9                                                        & -                                                                   & -                                                                           &  \ding{51}          &  \ding{55}                                                           & en       & Codeforces \\
LiveCodeBench\citep{DBLP:journals/corr/abs-2403-07974} & $\star$$\star$         & 511         & 271.5                                                        & 3.85M                                                               & -                                                                           &  \ding{55}           &  \ding{55}                                                           & en       & LeetCode   \\
EffiBench\citep{DBLP:conf/nips/0005QSCZ24}             & $\star$             & 1000        & 212                                                          & 82.5k                                                               & 16                                                                          &  \ding{55}           &  \ding{51}                                                          & en       & LeetCode   \\ \hline
OIBench                                                                       & $\star$$\star$$\star$$\star$ & 250         & 225.3                                                        & 37.5M                                                               & 75.2                                                                        &  \ding{51}          &  \ding{51}                                                          & zh/en    & Private    \\ \hline
\end{tabular}
\caption{Basic statistics, information between ours OIBench and other code benchmarks.
}
\label{tab/stat}

\end{table}

We present a comparative analysis between our benchmark and competing benchmarks in Table~\ref{tab/stat}. The difficulty levels are calibrated based on the accuracy rates of GPT-4O (as shown in Appendix Table~\ref{tab/diffComp}). Our OIBench significantly exceeds competing benchmarks in terms of problem difficulty, and test case size. Furthermore, we ensure that at the time of dataset release (May 2025), none of the problems can be directly retrieved from the Internet.

\subsection{Anti-Contamination}

A challenge for benchmark design is the in-distribution contamination~\cite{DBLP:journals/corr/data_agnostic_detect} problem, as modern LLMs typically scrape Internet data for pre-training and fine-tuning, potentially including contents with the same domain of the benchmark (e.g. the training set of GSM8k). The contamination not only renders leaked problems ineffective as test cases, but also artificially inflates performance scores on benchmarks of the same domain without actual improvements on the corresponding abilities.\citep{DBLP:journals/corr/sky-work,DBLP:journals/corr/gair_paper,DBLP:conf/naacl/Deng0TGC24,DBLP:journals/corr/abs-2406-04244}

To assess OIBench's robustness against the contamination, we conducted experiments using 100 held-out C++ problems with official solutions as training targets. We mix these problems with 10,000 standard supervised fine-tuning  samples to simulate data contamination scenarios. We train two variants: baseline models using only standard SFT data, and leaked models additionally trained on the 100 held-out problems.

We propose a Risk-Score metric to quantify how in-domain data might lead to contamination. The metric balances absolute and relative improvements by measuring progress toward perfect accuracy:
$
\text{Risk Score} = \frac{S_{\text{contaminated}} - S_{\text{baseline}}}{1 - S_{\text{baseline}}}
$, where $S_{\text{contaminated}}$ and $S_{\text{baseline}}$ represent scores for the respective model variants.



Table \ref{tab/risk_score} shows the Risk-Score of OIBench on different size models. We see that the Risk-Score is extremely low($<0.01$) for various models. The experiment shows that models cannot obtain improvements on OIBench unless they obtain comprehensive improvements on all abilities that are related to coding. For example, from Llama3-8B to Deepseek-V3.
We owe this robustness to the difficulty of OIBench, we believe that the presence of potential in-distribution data will not adversely affect OIBench's evaluation of the actual code capabilities of LLMs.

\begin{table}[h]
\tiny
\centering
\begin{tabular}{c|c c c c c c}
    \hline
     & Llama3-8B & Qwen2.5-7B & Qwen2.5-14B & Qwen2.5-72B & Qwen2.5-Coder-32B & Deepseek-V3 \\ \hline
    Risk-Score & 0.00 & 0.00 & 0.00 & 0.01 & 0.01 & 0.01 \\
    \hline
\end{tabular}

\caption{Risk-Score of OIBench on different models.}
\label{tab/risk_score}
\end{table}

\vspace{-5pt}

\section{OIBench Learderboards}

In this section, we report the evaluation results on current state-of-the-art LLMs. By default, we give all the information 

\textbf{Evaluation Environment.}
We evaluate all implementations on a computing server running CentOS Linux 7.6.1810 (Core) with Linux kernel x86-64. For each language:
For C++ implementations, we compile the code using g++ 11.3.0 with C++17 standard compliance and -O2 optimization flags.
For JavaScript implementations, we execute the code using Node.js v16.18.1 with npm v8.19.2 for dependency management.
For Java implementations, we run the programs using Java 1.8.0\_45 with the Java HotSpot 64-Bit Server VM.
For Python components, we employ Python 3.9 for orchestration.
The environment operates under Linux kernel 5.4.0 with glibc 2.31, using Docker~\citep{10.5555/2600239.2600241} containers with privileged hardware access. We monitor resources through system utilities (time command and /proc filesystem analysis) and configure environment variables for optimal runtime performance.

\textbf{Baseline Models.}
We mainly report code performance of 18 state-of-the-art LLMs, including 14 instruct models and 4 base models, the details for each model is elaborted in Table in Appendix. We list the prompts for instruct or base models respectively in Appendix. 
All models are evaluated in zero-shot, including the base models. For reasoning models, we set the maximum inference tokens to be 32,768.
For each model, we evaluate its coding ability on 4 languages, namely C++/Python/Java/JavaScript. 

By default, we use greedy search sampling with temperature as 0 for all tested models for output stability, except for the DeepSeek-R1 and Qwen3-32B model, where we find that its reasoning process tends to collapse in recursive patterns under greedy search. We set its temperature at 0.6 as its paper suggests~\citep{DBLP:journals/corr/abs-2501-12948}.  

\subsection{OIBench Leaderboard}
We list the evaluation results in Table~\ref{tab/main}, and report the confidence interval~\citep{miller2024adding} in appendix~\ref{sec:confidence}.

We find that reasoning models achieve an average Overall score of $21.4\%$ on OIBench, significantly outperforming instruction-tuned models, which score around $3.6\%$. Among all models, O4-mini-high ranks highest, consistently outperforming across all languages and tasks. This demonstrates the reasoning models' advantage in solving complex problems and highlights OIBench's ability to evaluate reasoning and chain-of-thought capabilities, distinguishing models in terms of reasoning ability.

Closed-source models score $14.5\%$ on average, while open-source models score $6.3\%$. This performance gap is consistent across both categories, as closed-source models generally have more compute resources and higher-quality training data~\citep{DBLP:conf/icml/EirasPVWPEMBCSB24}.

For instruction-tuned models, coding performance on OIBench strongly correlates with base model capabilities. Given the zero-shot evaluation, this suggests that base models already possess substantial coding potential, with instruction tuning providing marginal improvements on coding. We encourage focusing on base model capabilities rather than fine-tuned variants. The exception is DeepSeek-V3-0324, which surpass all other non-reasoning models. As DeepSeek-V3-0324 employs DeepSeek-R1’s CoT for distillation, which enhances the reasoning capabilities. We provide a detailed analysis of this improvement in later sections.

Regarding programming languages, models perform over $10\%$ worse on JavaScript and Python compared to C++ and Java on average. This discrepancy likely stems from factors such as skewed data distributions and training biases, which requires further investigation.
Regarding natural languages, models show minimal performance differences between Chinese and English, with Chinese performing slightly better, likely due to a latent source language preference in prompt translation.

\begin{table}[]
\centering
\scriptsize
\setlength{\tabcolsep}{2.5pt}
\begin{tabular}{c|ccccc|cc|ccccc|cc}
\hline
Model                           & \multicolumn{5}{c|}{OIBench}                                 & \multicolumn{2}{c|}{\begin{tabular}[c]{@{}c@{}}OIBench\\ by language\end{tabular}} & \multicolumn{5}{c|}{OIBench Pseudo}                          & \multicolumn{2}{c}{\begin{tabular}[c]{@{}c@{}}OIBench Pseudo\\ by language\end{tabular}} \\ \cline{2-15} 
                                & \multicolumn{1}{c|}{Overall} & C++   & Java  & Python & JS    & En                                       & Zh                                       & \multicolumn{1}{c|}{Overall} & C++   & Java  & Python & JS    & En                                          & Zh                                          \\ \hline
Base Models                     & \multicolumn{1}{c|}{}        &       &       &        &       &                                          &                                          & \multicolumn{1}{c|}{}        &       &       &        &       &                                             &                                             \\ \hline
DeepSeek-V3-base~\citep{liu2024deepseek}                & \multicolumn{1}{c|}{2.85}    & 5.20  & 4.20  & 2.00   & 0.00  & 2.30                                     & 3.40                                     & \multicolumn{1}{c|}{27.45}   & 41.60 & 37.00 & 29.40  & 1.80  & 28.20                                       & 26.70                                       \\
Qwen2.5-Coder-32B-Base~\citep{hui2024qwen2}          & \multicolumn{1}{c|}{0.95}    & 0.60  & 2.00  & 1.00   & 0.20  & 0.90                                     & 1.00                                     & \multicolumn{1}{c|}{17.80}   & 21.80 & 22.80 & 24.40  & 2.20  & 17.80                                       & 17.80                                       \\
Qwen2.5-72B-base~\citep{yang2024qwen2}                & \multicolumn{1}{c|}{0.75}    & 1.20  & 1.00  & 0.40   & 0.40  & 0.80                                     & 0.70                                     & \multicolumn{1}{c|}{18.50}   & 21.60 & 25.60 & 26.20  & 0.60  & 18.80                                       & 18.20                                       \\
Llama-3.1-405B-base~\citep{grattafiori2024llama}             & \multicolumn{1}{c|}{0.75}    & 1.20  & 1.40  & 0.40   & 0.00  & 0.60                                     & 0.90                                     & \multicolumn{1}{c|}{18.40}   & 19.80 & 25.60 & 28.00  & 0.20  & 18.90                                       & 17.90                                       \\ \hline
Instruction Tuned Models        & \multicolumn{1}{c|}{}        &       &       &        &       &                                          &                                          & \multicolumn{1}{c|}{}        &       &       &        &       &                                             &                                             \\ \hline
Qwen2.5-Coder-32B~\citep{hui2024qwen2}               & \multicolumn{1}{c|}{1.90}    & 1.00  & 2.40  & 2.80   & 1.40  & 1.40                                     & 2.40                                     & \multicolumn{1}{c|}{22.90}   & 20.80 & 26.40 & 30.40  & 14.00 & 21.30                                       & 24.50                                       \\
Qwen2.5-72B~\cite{yang2024qwen2}                     & \multicolumn{1}{c|}{1.80}    & 1.20  & 2.80  & 2.40   & 0.80  & 1.50                                     & 2.10                                     & \multicolumn{1}{c|}{19.00}   & 20.40 & 24.40 & 24.20  & 7.00  & 18.90                                       & 19.10                                       \\
Llama-3.1-405B~\citep{grattafiori2024llama}                  & \multicolumn{1}{c|}{1.15}    & 1.60  & 1.00  & 1.20   & 0.80  & 1.40                                     & 0.90                                     & \multicolumn{1}{c|}{28.45}   & 28.80 & 32.00 & 35.00  & 18.00 & 29.90                                       & 27.00                                       \\
DeepSeek-V3-1226~\citep{liu2024deepseek}                & \multicolumn{1}{c|}{4.10}    & 4.80  & 4.80  & 3.80   & 3.00  & 3.90                                     & 4.30                                     & \multicolumn{1}{c|}{32.67}   & 38.00 & 36.80 & 31.90  & 24.00 & 32.95                                       & 32.40                                       \\
DeepSeek-V3-0324~\citep{liu2024deepseek}                & \multicolumn{1}{c|}{11.60}   & 13.20 & 11.60 & 13.80  & 7.80  & 11.30                                    & 11.90                                    & \multicolumn{1}{c|}{32.15}   & 35.00 & 36.80 & 29.80  & 27.00 & 32.40                                       & 31.90                                       \\
Doubao-32k-pro~\citep{doubao2025doubao}$^\text{\faLock}$ & \multicolumn{1}{c|}{2.30}    & 1.40  & 4.00  & 2.40   & 1.40  & 1.90                                     & 2.70                                     & \multicolumn{1}{c|}{22.50}   & 27.60 & 14.00 & 30.60  & 17.80 & 20.30                                       & 24.70                                       \\
GPT-4O~\citep{hurst2024gpt}$^\text{\faLock}$         & \multicolumn{1}{c|}{2.65}    & 1.80  & 4.40  & 2.80   & 1.60  & 2.70                                     & 2.60                                     & \multicolumn{1}{c|}{21.45}   & 21.00 & 23.80 & 23.40  & 17.60 & 19.90                                       & 23.00                                       \\
Claude3.5 Sonnet~\citep{claude}$^\text{\faLock}$      & \multicolumn{1}{c|}{3.45}    & 2.80  & 6.20  & 2.20   & 2.60  & 3.60                                     & 3.30                                     & \multicolumn{1}{c|}{28.90}   & 37.00 & 26.40 & 27.00  & 25.20 & 25.80                                       & 32.00                                       \\ \hline
Reasoning Models                & \multicolumn{1}{c|}{}        &       &       &        &       &                                          &                                          & \multicolumn{1}{c|}{}        &       &       &        &       &                                             &                                             \\ \hline
QwQ-32B~\citep{qwq32b}                         & \multicolumn{1}{c|}{10.65}   & 10.60 & 13.60 & 8.40   & 10.00 & 10.30                                    & 11.00                                    & \multicolumn{1}{c|}{31.70}   & 29.60 & 33.00 & 39.40  & 24.80 & 33.40                                       & 30.00                                       \\
DeepSeek-R1~\citep{DBLP:journals/corr/abs-2501-12948}                     & \multicolumn{1}{c|}{20.50}   & 22.40 & 23.20 & 21.40  & 15.00 & 19.80                                    & 21.20                                    & \multicolumn{1}{c|}{39.15}   & 46.20 & 38.00 & 39.20  & 33.20 & 37.50                                       & 40.80                                       \\
Qwen3-32B~\citep{qwq32b}                       & \multicolumn{1}{c|}{18.95}   & 20.60 & 21.40 & 19.20  & 14.60 & 20.40                                    & 17.50                                    & \multicolumn{1}{c|}{33.35}   & 34.40 & 35.40 & 35.20  & 28.40 & 35.60                                       & 31.10                                       \\
O1~\citep{jaech2024openai}$^\text{\faLock}$             & \multicolumn{1}{c|}{15.40}   & 22.00 & 21.20 & 6.60   & 11.80 & 15.70                                    & 15.10                                    & \multicolumn{1}{c|}{32.25}   & 44.80 & 36.40 & 18.20  & 29.60 & 30.70                                       & 33.80                                       \\
O3-mini-high~\citep{O3}$^\text{\faLock}$   & \multicolumn{1}{c|}{26.80}   & 31.80 & 26.80 & 28.40  & 20.20 & 27.30                                    & 26.30                                    & \multicolumn{1}{c|}{44.90}   & 54.60 & 44.00 & 42.00  & 39.00 & 45.20                                       & 44.60                                       \\
O4-mini-high~\citep{O4}$^\text{\faLock}$   & \multicolumn{1}{c|}{36.35}   & 43.40 & 37.20 & 34.60  & 30.20 & 35.60                                    & 37.10                                    & \multicolumn{1}{c|}{51.70}   & 58.80 & 53.40 & 49.80  & 44.80 & 52.80                                       & 50.60                                       \\ \hline
\end{tabular}

\caption{Main Leaderboard in accepted rate. We adopt the AC (All Correct) rate (\%) in this table, where a model must pass all test cases of a problem to be considered correct. 
 $\text{\faLock}$ indicates close-source models. }
\label{tab/main}
\end{table}

\subsection{OIBench with Pseudocode}
As the previous section reports, our proposed OIBench is highly complex for non-reasoning models and base models to obtain correct answers without reasoning process. In real-world scenarios,  a finer-grained analysis of LLM performance on code; simultaneously, we want to evaluate models' understanding of the reasoning process. Therefore, we employ R1 to convert canonical solutions into pseudocode and use it as prompts. We ensure that the pseudocode remains language-agnostic. When evaluating problems, we provide both the problem statement and the pseudocode as prompts to test the models' problem-solving capabilities, we provide results in Table~\ref{tab/main}.

With the help of pseudocode solutions, all models show a significant performance improvement when provided with solution hints. Even for the strongest models, such as O3-mini-high and O4-mini-high, OIBench Pseudo results in a noticeable boost in performance. This indicates that the solution hints greatly reduce the difficulty that originally relied on complex reasoning, narrowing the gap between Reasoning Models and Instruction Tuned Models. This suggests that this evaluation method is more suitable for assessing models' ability to understand solutions and their code generation capabilities. At the same time, Reasoning Models still maintain an advantage in solution comprehension.

Furthermore, we observe that the performance of Instruction Tuned Models remains highly correlated with that of their corresponding Base models, further indicating that the code generation potential of a model is closely tied to the level of its pretraining.


\subsection{Code Efficiency Metrics}

OI competitions' test cases are designed to be large-scale, ensuring comprehensive coverage of edge cases and providing rigorous evaluation of algorithmic time/space complexity. OIBench follows this standard with large-scaled test cases.
Conventional benchmarks typically employed time/space thresholds as acceptance criteria, which suffer from two limitations: (1) they only coarsely approximate whether solutions meet specific complexity classes, (2) threshold selection can be subjective, where different threshold leads to performance fluctuation. 

EffiBench~\citep{DBLP:conf/nips/0005QSCZ24} introduced efficiency metrics based on runtime/space ratios to canonical solutions, however, this metrics remain insufficiently intuitive: (1) it reports a single numerical value that fails to capture temporal distribution of different LLMs, (2) the metric becomes unreliable when measuring low-complexity but incorrect solutions, which can be frequent in LLMs' generation.

\begin{figure}[h]
    \centering
    \begin{subfigure}[b]{0.48\textwidth}
        \centering
        \includegraphics[width=\textwidth]{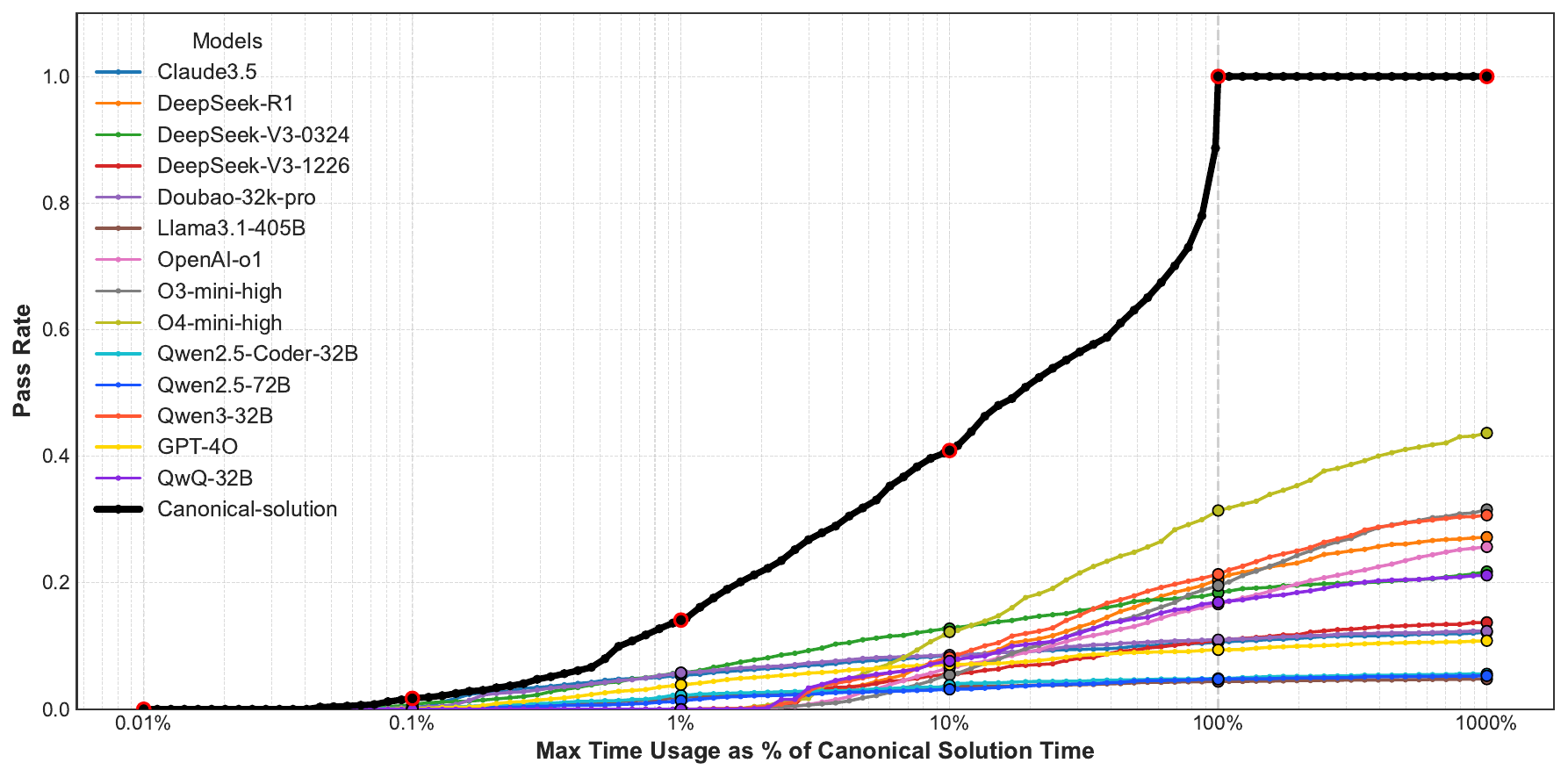}
        \caption{Time Completion Curve}
        \label{fig:sub1}
    \end{subfigure}
    \hfill
    \begin{subfigure}[b]{0.48\textwidth}
        \centering
        \includegraphics[width=\textwidth]{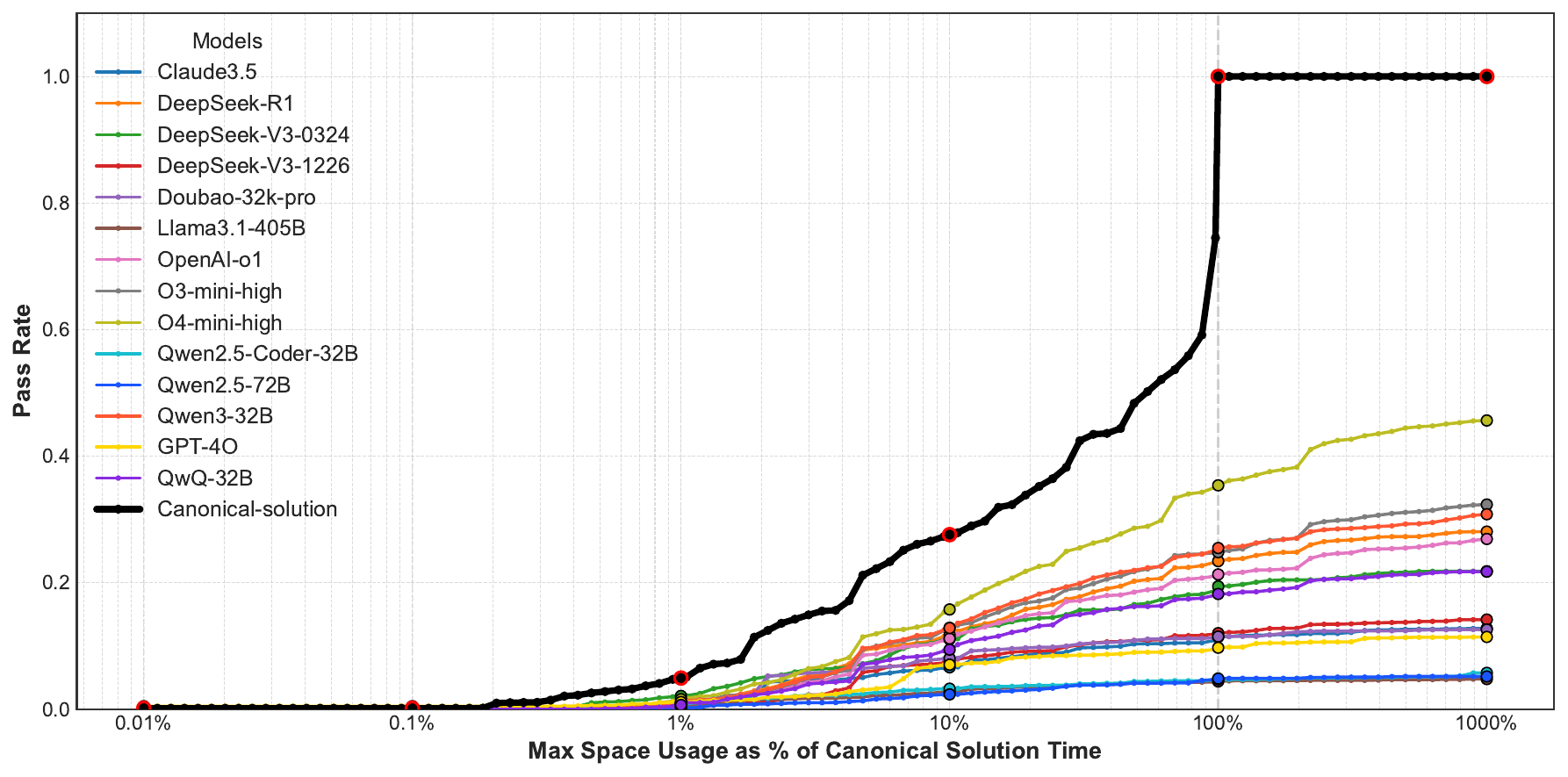}
        \caption{Space Completion Curve}
        \label{fig:sub2}
    \end{subfigure}
    \caption{ Time/Space Completion Curves. Note that in the figure, we employ the test case pass rate rather than the AC rate, as the former provides more fine-grained visualization.}
    \label{fig:timespace}
\end{figure}

Our paper presents a fine-grand efficiency evaluation metric, namely the \textbf{Time/Space-Completion Curve}, as shown in Figure~\ref{fig:timespace}. This curve is essentially the cumulative distribution of solving time/space, offering significant advantages, where the X-axis represents time/space threshold (in ratio to the canonical solution) on log-scale, while the Y-axis is the fraction of solved test cases under the threshold.

The analysis reveals several key insights:
\begin{enumerate}[left=0pt]
\item  The models' algorithms underperform the canonical solution in both time and space complexity. Specifically, even after reaching 100\% of canonical usage, all models still exhibit growth, suggesting potential for further inprovment in algorithmic efficiency.

\item  O4-mini-high generally demonstrates the best time and space optimization. However, in the low time region (less than 10\% of canonical usage), it performs worse than models such as GPT-4O and DeepSeek-V3. This suggests that the algorithmic design of non-reasoning models may perform better on simpler test cases, as they have learned plenty naive and inefficient algorithms.

\item  Additionally, non-reasoning models are generally inefficient in their algorithmic design, while reasoning models exhibit a significant advantage in optimizing time and space complexity. This underscores the strength of reasoning models in handling complex tasks more efficiently than instruction-tuned models.
\end{enumerate}

\vspace{-14pt}

\subsection{Test-Time Compute Comparison}

\begin{figure}[htbp]
    \centering
    \includegraphics[width=0.7\linewidth]{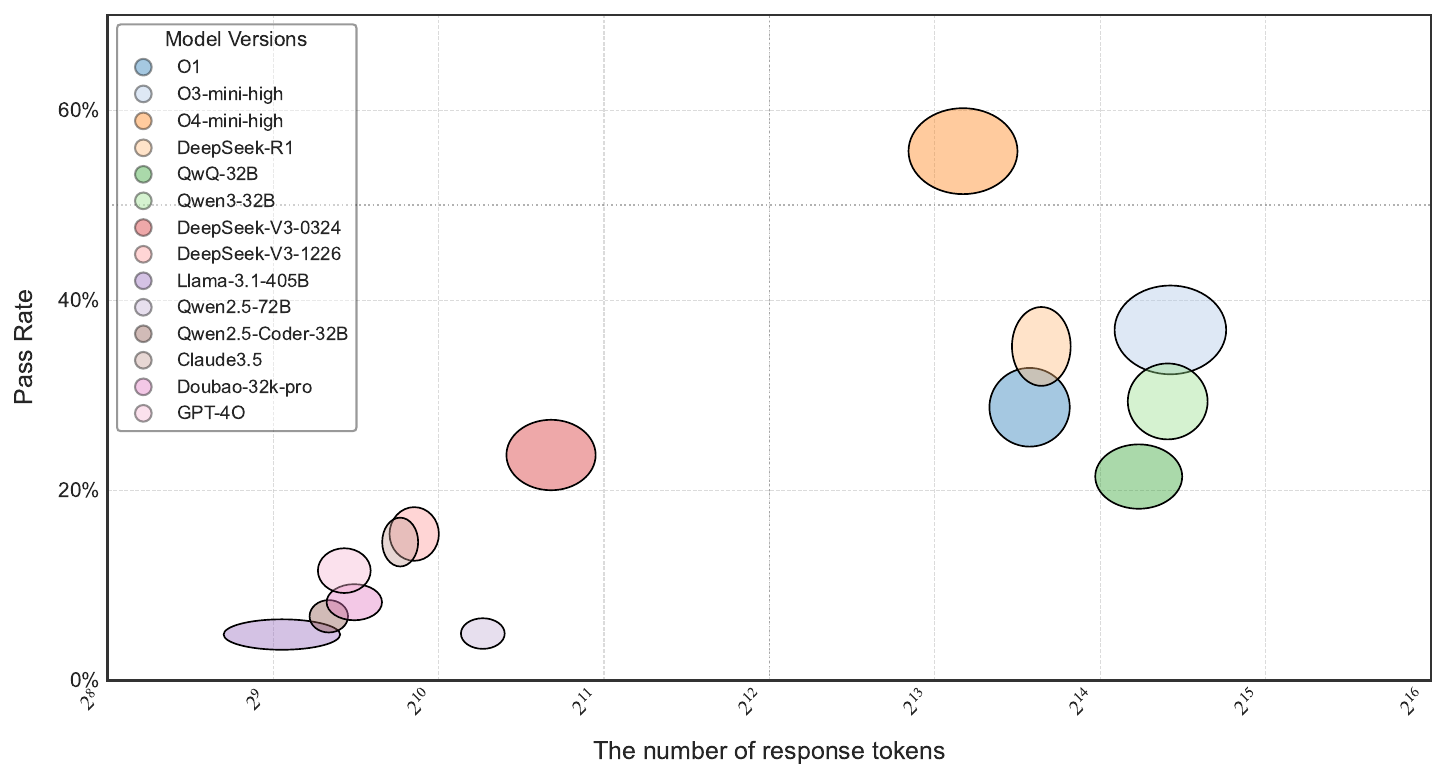}
    \caption{Inference budgets measured by the number of response tokens vs pass rate. The size of the bubble indicates the standard deviations on the corresponding axes.}
    \label{fig:inferBudgets}
\end{figure}

Since test-time computation has become a paradigm to improve LLMs' performance on complex tasks~\cite{DBLP:journals/corr/abs-2408-03314}, recent models like OpenAI-O1 and DeepSeek-R1 utilize long Chain-of-Thought as the main test-time scaling technique. 
Given that similar methods incur substantial computational budget during inference, researchers typically examine the trade-off between model computational cost and performance. We report the distribution of model performance versus the number of inference tokens\footnote{For close-sourced models, we report the token usage from its API return value.} in Figure~\ref{fig:inferBudgets}.  
O4-mini-high demonstrates the best reasoning efficiency, solving more problems correctly with fewer tokens. Qwen3-32B and QwQ-32B exhibit similar pass rates to DeepSeek-R1 and O1, respectively, but require significantly more tokens for reasoning. Among the non-inference models, DeepSeek-V3-0324 achieves the best performance while also having the longest reasoning token count, suggesting that it has learned certain characteristics of CoT long-chain reasoning during the distillation learning of R1.

\vspace{-18pt}

\section{Comprehensive Analysis}

\subsection{Human-comparable Evaluation}
\label{sec:Human}
CodeElo~\citep{DBLP:journals/corr/abs-2501-01257} was the first study that provided an LLM-human performance comparison in code via an online code competition site CodeForces. However, online competition benchmarks exhibits certain limitations. 
Their use of human Elo ratings from contests up to \textbf{six months} prior introduces a temporal misalignment, potentially leading to outdated comparisons, and difficulty in reproducibility. Additionally, their construct the Elo system via randomly sampling 250 user scores from the Codeforces website, leading to imprecision.

\begin{figure}[htbp]
    \centering
    \includegraphics[width=0.7\linewidth]{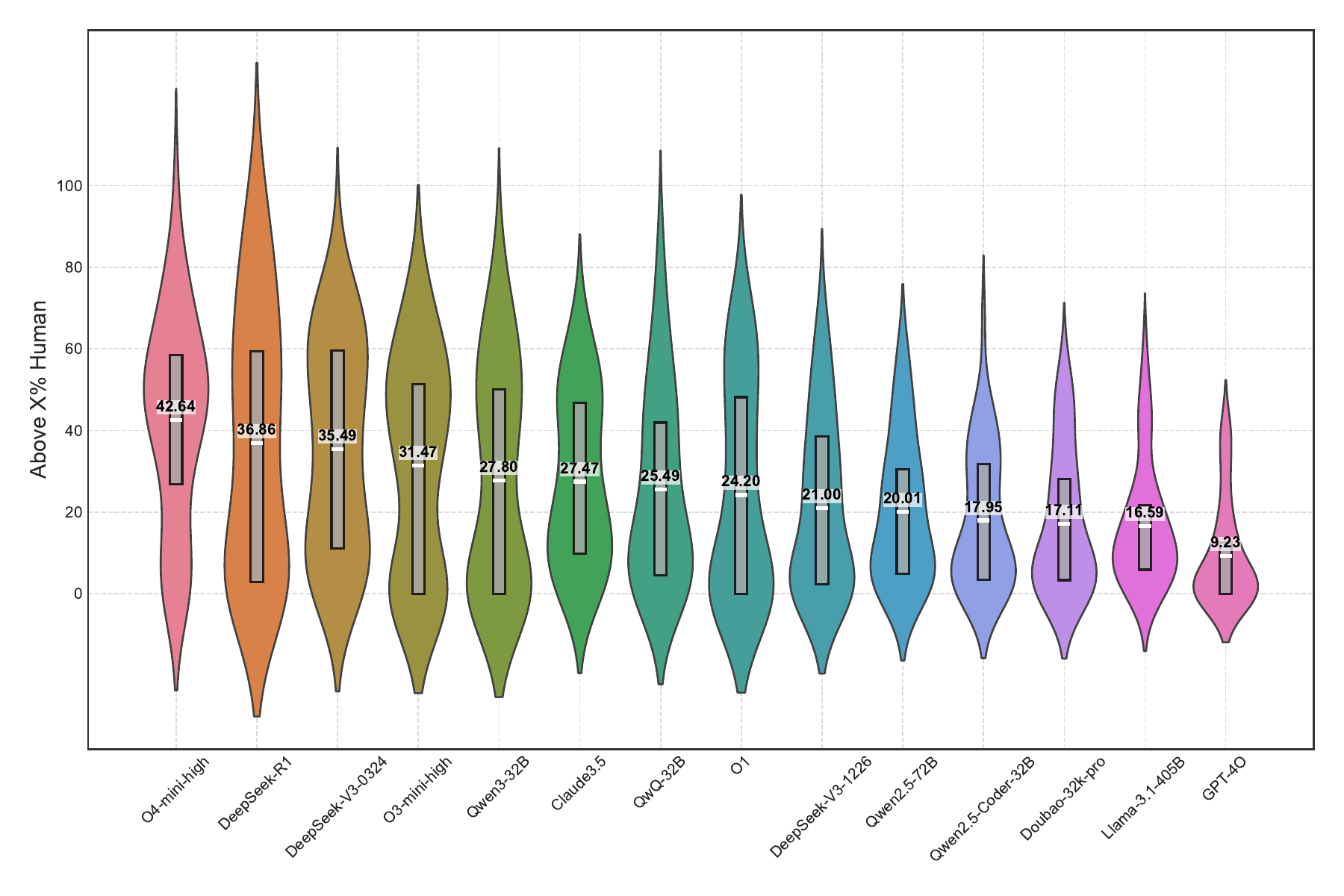}
    \caption{Violin plots of model's relative ranking among human participants. The number indicates the average relative ranking for each model.}
    \label{fig:distributionVShuman}
\end{figure}

OIBench further conducts a systematic comparison between LLMs and human programmers via introducing static human performance metrics. Specifically, we curate a set of 44 programming problems from OIBench and recruit ACM-ICPC contestants to complete them under controlled conditions, subsequently reporting aggregated results. To enhance reproducibility and facilitate deeper analysis, we release the anonymized solution records alongside the benchmark dataset, enabling the research community to rigorously examine the performance disparities between humans and large language models in code generation tasks. For each model, we report its ranking among all human participants via violin plots, as shown in Fig~\ref{fig:distributionVShuman}.

Our ranking follows the IOI competition rules: Each submission is first evaluated on the basis of the number of passed test cases, with higher numbers resulting in better rankings. When submissions have the same number of test cases passed, the ranking is determined by execution time, where shorter times lead to higher rankings. For human participants, we consider their last submission as their final answer for comparison.

\vspace{-5pt}

We observe that the violin plots of the models’ relative ranking among human participants reveal three distinct performance patterns:

\begin{enumerate}[left=0pt]
    \item Bass-shaped: These models tend to rank consistently low across most problems, typically outperforming less than 20\% of human participants. All models in this category are large language models (LLMs) without long-chain-of-thought (long CoT) capabilities. In practice, these models can only solve a limited number of test cases and fail to reach competitive problem-solving levels.
    \item Peanut-shaped : These models perform well on certain problems (e.g., surpassing around 50\% of human participants), but poorly on others. Most models in this category support long CoT reasoning, with the exception of DeepSeek-V3-0324. Their violin plots consistently exhibit a bimodal distribution, which may reflect the underlying train-inference paradigm. The presence of two peaks suggests that current long CoT models are able to solve certain types of problems effectively, while still lacking the reasoning capabilities required for other types.
    \item Olive-shaped: This pattern features a more uniform distribution of rankings across tasks. At present, only O4-mini-high exhibits this shape. Intuitively, such a distribution more closely resembles the performance spread observed in human populations.
\end{enumerate}

The contrast between the peanut-shaped and olive-shaped distributions is particularly noteworthy. The peanut-shaped trend highlights a potential limitation of many current long CoT training paradigms—models tend to learn strategies effective for only a subset of problem types, a topic we discuss in further detail later. In contrast, the reasoning capability of O4-mini appears more human-like in its generality and consistency.

We strongly advocate for leaderboard to include model-versus-human rating comparisons. Such visual and statistical assessments help the research community develop a more nuanced understanding of model generalization and alignment with human capabilities.

\subsection{Qualitative Analysis}
\label{sec:QA}
We perform qualitative analysis of error patterns across multiple models, categorizing algorithmic erros or flaws. With assistance from DeepSeek-R1 and validation by an ACM-ICPC-level programmer, we annotate and visualize error distributions of serveral models in appendix Figure~\ref{figure/error_distribution_final}.

For most reasoning models, the primary cause of failure is that the Chain-of-Thought (CoT) becomes too long and cannot complete reasoning within the limited token constraint. Given our relatively large maximum reasoning length setting (32,768), we observe that these failed long reasoning chains exhibit noticeable recursive patterns. We suggest that future improvements in reasoning efficiency for CoT and mitigation of recursive  patterns will be crucial for enhancing models' reasoning capabilities, as our previous analysis has shown that O4-mini-high achieves the highest CoT reasoning efficiency and obtains superior results. Additionally, incorrect algorithm implementation and inefficient algorithm selection remain major obstacles affecting model performance in reasoning tasks. While issues such as program compilation errors and data processing errors occur less frequently, they still exist.

\section{Conclusion}

This paper introduces OIBench, a verified, private, and challenging benchmark designed to evaluate LLMs on Olympiad-level algorithmic problems. Our contributions are as follows.

First, OIBench is highly challenging even for frontier reasoning models like O4-mini-high. Existing benchmarks increasingly lose the ability to differentiate models' reasoning abilities due to saturation problems. In contrast, reasoning models substantially outperform non-reasoning models on OIBench, highlighting the benchmark's strong discriminative power. 

Secondly, our Time/Space Completion Curves based on OIBench reveal considerable room for improvement in the reasoning efficiency of current models, particularly through optimizing algorithms in terms of time and space complexity.

Thirdly, human-model comparisons on OIBench show that a gap remains between LLMs and top-tier human players. Though frontier reasoning models surpass the majority of ACM-level competitors, they still lag far behind top-tier human players. However, O4-mini-high exhibits a distinct pattern compared to other models. We encourage further research into human-comparable analysis to gain deeper insights in this light.

Last but not least, for reasoning models, we suggest that optimizing the efficiency of Chain-of-Thought is a crucial direction for future advancements in reasoning capabilities.
By releasing OIBench as a fully open-source resource benchmark (including problems, test cases, and canonical solutions), we aim to help advance the development of LLMs' reasoning abilities.

\clearpage
\newpage

\appendix

\newpage

\section{Appendix}

\subsection{Limitations and Further Studies}
\label{sec:limitation}

Despite rigorous data cleaning procedures and verification through search engines, the dataset may still carry potential risks of internet leakage. Additionally, we cannot completely ensure the absolute originality of the coaching questions in terms of problems' paradigms, as some might be adapted from existing problem types.

The observed performance disparity between models implemented in C++/Java and those in Python/JS remains unexplained and warrants further exploration from perspective of model training.

Our anti-contamination experiments excluded reinforcement learning (RL) training components. Standard practice dictates that internet-sourced pre-training data is typically excluded from RL training phases. However, emerging methodologies like pretraining with RL~\citep{wei2025swe} could potentially introduce contamination risks following OIBench's open-source release.

\subsection{Difficulty Comparsion}

We list the competitive benchmarks' results in Table~\ref{tab/diffComp} with GPT-4O as the evaluated model. 
We report the HumanEval's result from \url{evalplus.github.io/leaderboard.html}. We use pass@8 results for CodeElo as their paper reported. We report the CodeContests's result from \cite{DBLP:journals/corr/abs-2411-14503}. For other benchmarks, we report the resultes from the corresponding paper.

\begin{table}[h]
\centering
\small
\setlength{\tabcolsep}{3pt}
\begin{tabular}{c|cccccc|c}
\hline
       & HumanEval & CodeContests & USACO & CodeElo & LiveCodeBench & EffiBench & OIBench \\ \hline
GPT-4O & 87.2    & 34.7         & 8.3   & 16.8   & 41.9          & 50.8      & 2.6     \\ \hline
\end{tabular}
\caption{Comparsion on between ours OIBench and other code benchmarks on difficulty, with GPT-4O as standard model. }
\label{tab/diffComp}
\end{table}

\subsection{Training details for anti-contamination experiments}
\label{sec:training}

We train the models using $10,000$ supervised fine-tuning (SFT) samples, mixed with $100$ samples from OIBench. For all models, we adopt the Megatron-LM~\citep{shoeybi2019megatron} framework for training, a learning rate ranging from $5 \times 10^{-6}$ to $8 \times 10^{-6}$, the Adam optimizer, and train for 3 epochs. The choice of these hyperparameters follows common practices in the open-source community for models with similar parameter and data scales, without additional hyperparameter search or special optimization.

\subsection{Derived Benchmarks}
To further evaluate LLMs' code comprehension abilities under more complex scenarios, and examine how well the model understands the complex logic of competition-level code,
we introduce four derived benchmarking tasks focusing on code understanding, which consists of the following subtasks:

\textbf{Bugfix(BF)}: Correct buggy solutions by identifying and fixing errors, similar to \citep{DBLP:journals/corr/abs-2411-02310}.

\textbf{Complete(CP)}: Complete partial solutions with missing code segments, similar to \citep{DBLP:conf/nips/LuGRHSBCDJTLZSZ21}.

\textbf{Translate(TS)}: Translate C++ solutions to equivalent Python implementations, similar to \citep{DBLP:conf/emnlp/YanTLCW23}.

\textbf{Interpret(IT)}: Interpret and predict outputs based on given solutions and test inputs, similar to \citep{DBLP:conf/icml/GuRLSS024}.

We list the results of the derived benchmark in Table~\ref{tab/derived}. 
Note that base models usually fail to generate valid results for comprehension tasks, therefore we exclusively evaluate the code understanding performance of instruction-tuned models.

The derived benchmark demonstrates similar overall trends to both OIBench and OIBench Pseudo, with reasoning models consistently outperforming non-reasoning models. Notably, models exhibiting strong performance on OIBench Pseudo also tend to display superior code comprehension capabilities. 
A particularly distinctive case is the code interpretation task, which shows greater discriminative power than other derived benchmarks and more effectively tests models' ability to simulate code execution through reasoning.

\begin{table}[]
\centering
\scriptsize
\setlength{\tabcolsep}{2.5pt}
\begin{tabular}{c|ccccc|cc}
\hline
Model                                                                          & \multicolumn{5}{c|}{OIBench Derived}                                                                         & \multicolumn{2}{c}{\begin{tabular}[c]{@{}c@{}}OIBench Derived\\ by language\end{tabular}} \\ \cline{2-8} 
                                                                               & Overall & \multicolumn{1}{l}{BF} & \multicolumn{1}{l}{CP} & \multicolumn{1}{l}{TS} & \multicolumn{1}{l|}{IT} & En                                          & Zh                                          \\ \hline
Instruction Tuned Models                                                       &         &                        &                        &                        &                         &                                             &                                             \\ \hline
Qwen2.5-Coder-32B~\citep{hui2024qwen2}                   & 15.45   & 21.60                  & 17.20                  & 8.60                   & 14.40                   & 16.60                                       & 14.30                                       \\
Qwen2.5-72B~\cite{yang2024qwen2}                         & 18.25   & 20.60                  & 18.80                  & 8.60                   & 25.00                   & 17.70                                       & 18.80                                       \\
Llama-3.1-405B~\citep{grattafiori2024llama}              & 21.00   & 28.80                  & 31.60                  & 12.40                  & 11.20                   & 25.00                                       & 17.00                                       \\
DeepSeek-V3-1226~\citep{liu2024deepseek}                 & 28.20   & 39.60                  & 35.80                  & 16.40                  & 21.00                   & 30.30                                       & 26.10                                       \\
DeepSeek-V3-0324~\citep{liu2024deepseek}                 & 33.25   & 34.60                  & 36.20                  & 21.60                  & 40.60                   & 34.80                                       & 31.70                                       \\
Doubao-32k-pro~\citep{doubao2025doubao}$^\text{\faLock}$ & 22.45   & 30.40                  & 26.40                  & 13.40                  & 19.60                   & 23.60                                       & 21.30                                       \\
GPT-4O~\citep{hurst2024gpt}$^\text{\faLock}$             & 27.55   & 39.00                  & 30.20                  & 15.80                  & 25.20                   & 29.90                                       & 25.20                                       \\
Claude3.5 Sonnet~\citep{claude}$^\text{\faLock}$         & 29.55   & 39.20                  & 38.80                  & 20.40                  & 19.80                   & 32.80                                       & 26.30                                       \\ \hline
Reasoning Models                                                               &         &                        &                        &                        &                         &                                             &                                             \\ \hline
QwQ-32B~\citep{qwq32b}                                   & 34.35   & 32.60                  & 25.60                  & 23.60                  & 55.60                   & 38.60                                       & 30.10                                       \\
DeepSeek-R1~\citep{DBLP:journals/corr/abs-2501-12948}    & 39.75   & 39.80                  & 43.60                  & 45.80                  & 29.80                   & 50.10                                       & 29.40                                       \\
Qwen3-32B~\citep{qwq32b}                                 & 28.30   & 28.20                  & 21.80                  & 20.40                  & 42.80                   & 33.70                                       & 22.90                                       \\
O1~\citep{jaech2024openai}$^\text{\faLock}$              & 40.70   & 36.60                  & 39.00                  & 19.80                  & 67.40                   & 40.00                                       & 41.40                                       \\
O3-mini-high~\citep{O3}$^\text{\faLock}$                 & 55.10   & 61.40                  & 49.00                  & 30.80                  & 79.20                   & 61.60                                       & 48.60                                       \\
O4-mini-high~\citep{O4}$^\text{\faLock}$                 & 59.75   & 62.60                  & 59.40                  & 31.40                  & 85.60                   & 66.80                                       & 52.70                                       \\ \hline
\end{tabular}
\caption{OIBench Derived Leaderboard in accepted rate. BF, CP, TS and IT stands for Bugfix, Code Complete, Code Translate and Code Interpret respectively. }
\label{tab/derived}
\end{table}

\subsection{Statistical Significance Report}
\label{sec:confidence}
We report confidence interval in Table~\ref{tab/confidence}, using the statistical approach from \citep{miller2024adding}.
While the confidence intervals of similar models overlap, our paper does not focus on ranking closely rival models. Instead, we highlight the significant performance gaps between different model tiers (e.g., reasoning models vs. conventional instruction-tuned models) on the Leaderboard, where the differences far exceed confidence intervals. This distinction is further supported by our extensive experiments.

\subsection{Computing Resources}
For anti-contamination experiments, models with fewer than $14$B parameters are trained on $8$ A100 GPUs ($80$GB each), which takes around $2$ hours; models with $32$B to $72$B parameters are trained on $8$ nodes, each equipped with $8$ A100 GPUs, which takes around $5$ hours. The Deepseek-V3 model is trained on $16$ nodes, each equipped with $8$ H20 GPUs ($141$GB each), which takes around $5$ hours.

For model evaluation on OIBench, we list the usage in Table~\ref{tab/confidence}. We use vLLM~\citep{kwon2023efficient} as the inference service. For close-sourced models, the evaluation time will fluctuate with the network and supplier services, so we do not provide a prediction time.

\begin{table}[h]
\centering
\tiny

\begin{tabular}{c|cc|cc}
\hline
                                & \multicolumn{2}{c|}{Confidence Interval} & \multicolumn{2}{c}{Computing Resources Usage}                                                \\ \cline{2-5} 
\multicolumn{1}{l|}{}           & OIBench         & OIBench Pseudo         & GPU type      & \begin{tabular}[c]{@{}c@{}}expected hours\\ for all experiments\end{tabular} \\ \hline
Base Models                     &                 &                        &               &                                                                              \\ \hline
DeepSeek-V3-base                & 3.01            & 3.08                   & H800-80G * 16 & 4                                                                            \\
Qwen2.5-Coder-32B-Base          & 1.14            & 2.89                   & A100-80G * 8  & 3                                                                            \\
Qwen2.5-72B-base                & 2.01            & 2.91                   & A100-80G * 16 & 4                                                                            \\
Llama-3.1-405B-base             & 2.01            & 2.91                   & A100-80G * 16 & 7                                                                            \\ \hline
Instruction Tuned Models        &                 &                        &               &                                                                              \\ \hline
Qwen2.5-Coder-32B               & 0.92            & 2.58                   & A100-80G * 8  & 2                                                                            \\
Qwen2.5-72B                     & 0.62            & 2.87                   & A100-80G * 8  & 3                                                                            \\
Llama-3.1-405B                  & 0.48            & 2.28                   & A100-80G * 16 & 7                                                                            \\
DeepSeek-V3-1226                & 1.01            & 2.57                   & H800-80G * 16 & 6                                                                            \\
DeepSeek-V3-0324                & 2.20            & 2.88                   & H800-80G * 16 & 8                                                                            \\
Doubao-32k-pro$^\text{\faLock}$ & 2.74            & 3.07                   & API           & -                                                                            \\
GPT-4O$^\text{\faLock}$         & 1.98            & 2.84                   & API           & -                                                                            \\
Claude3.5$^\text{\faLock}$      & 0.87            & 2.64                   & API           & -                                                                            \\ \hline
Reasoning Models                &                 &                        &               &                                                                              \\ \hline
QwQ-32B                         & 0.91            & 2.42                   & H800-80G * 16 & 18                                                                           \\
DeepSeek-R1                     & 0.55            & 2.31                   & H800-80G * 16 & 24                                                                           \\
Qwen3-32B                       & 1.24            & 2.53                   & H800-80G * 16 & 20                                                                           \\
O1$^\text{\faLock}$             & 1.25            & 2.89                   & API           & -                                                                            \\
O3-mini-high$^\text{\faLock}$   & 2.53            & 3.03                   & API           & -                                                                            \\
O4-mini-high$^\text{\faLock}$   & 1.80            & 2.49                   & API           & -                                                                            \\ \hline
\end{tabular}

\caption{Confidence interval and computing resources usage for each model. }
\label{tab/confidence}
\end{table}

\subsection{Time/Space-Completion Curve vs Human}

\begin{figure}[h]
    \centering
    \begin{subfigure}[b]{0.48\textwidth}
        \centering
        \includegraphics[width=\textwidth]{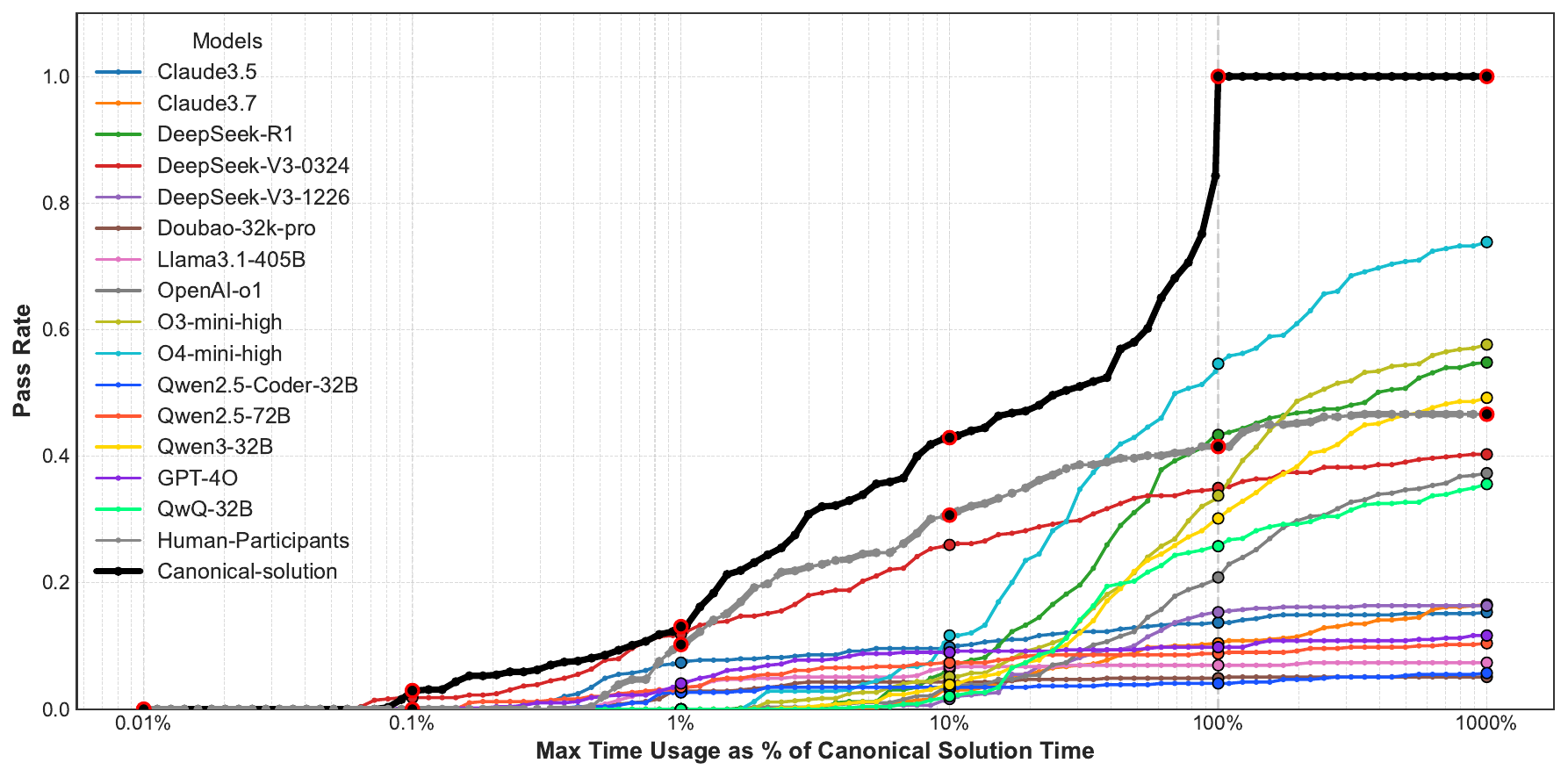}
        \caption{Time Completion Curve Compare with Human}
    \end{subfigure}
    \hfill
    \begin{subfigure}[b]{0.48\textwidth}
        \centering
        \includegraphics[width=\textwidth]{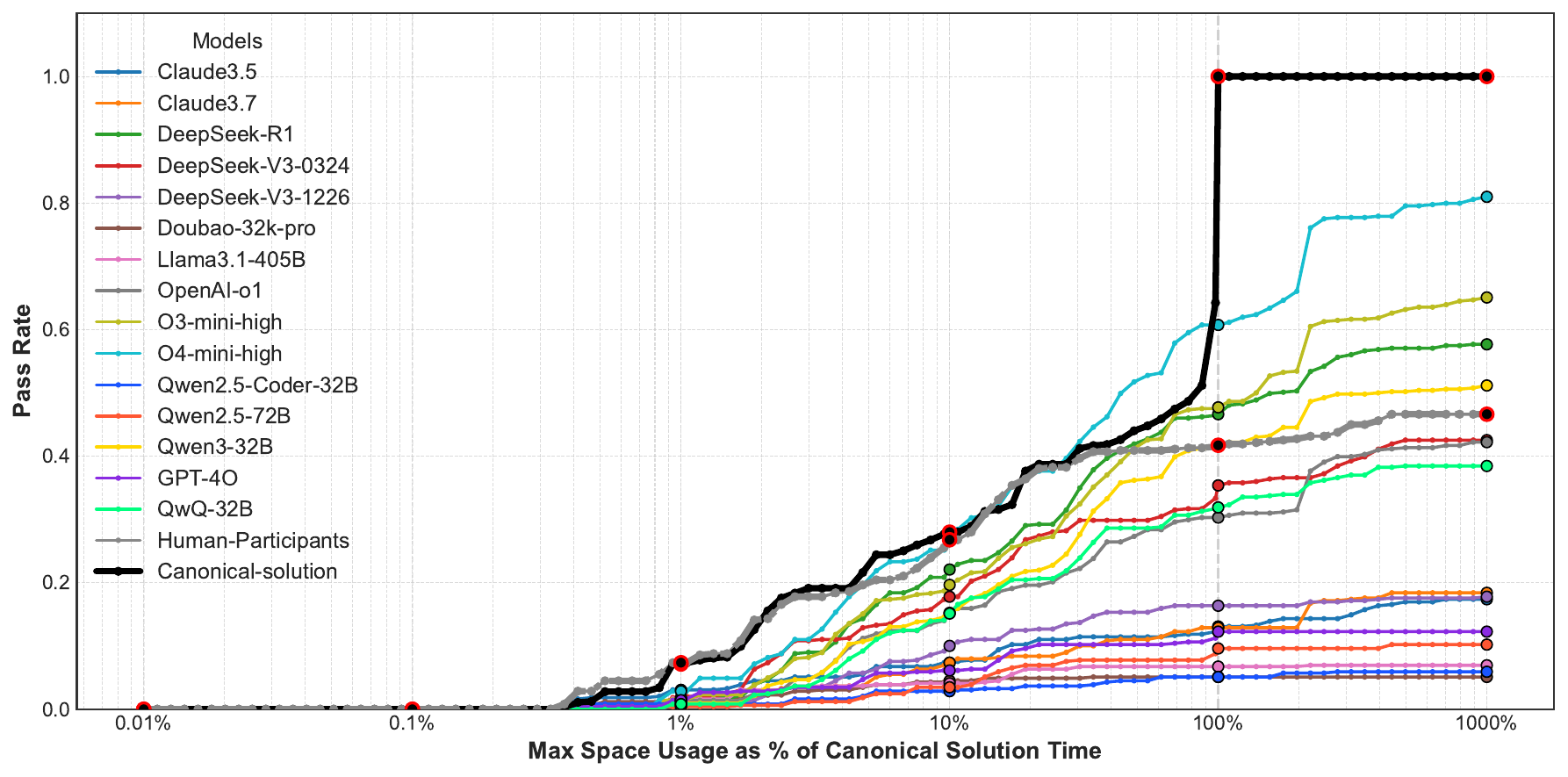}
        \caption{Space Completion Curve Compare with Human}
    \end{subfigure}
    \caption{ Time/Space Completion Curves}
    \label{fig:comHumanTS}
\end{figure}

We compare the models' performance with human contestants' in Figure~\ref{fig:comHumanTS} on 44 problems. 
We can see the SOTA reasoning models' solution efficiency surpass most human. The detailed analysis shown in Section~\ref{sec:Human}.



\subsection{Error Type Attribution}

\begin{figure}[htbp]
    \centering
    \includegraphics[width=0.8\linewidth]{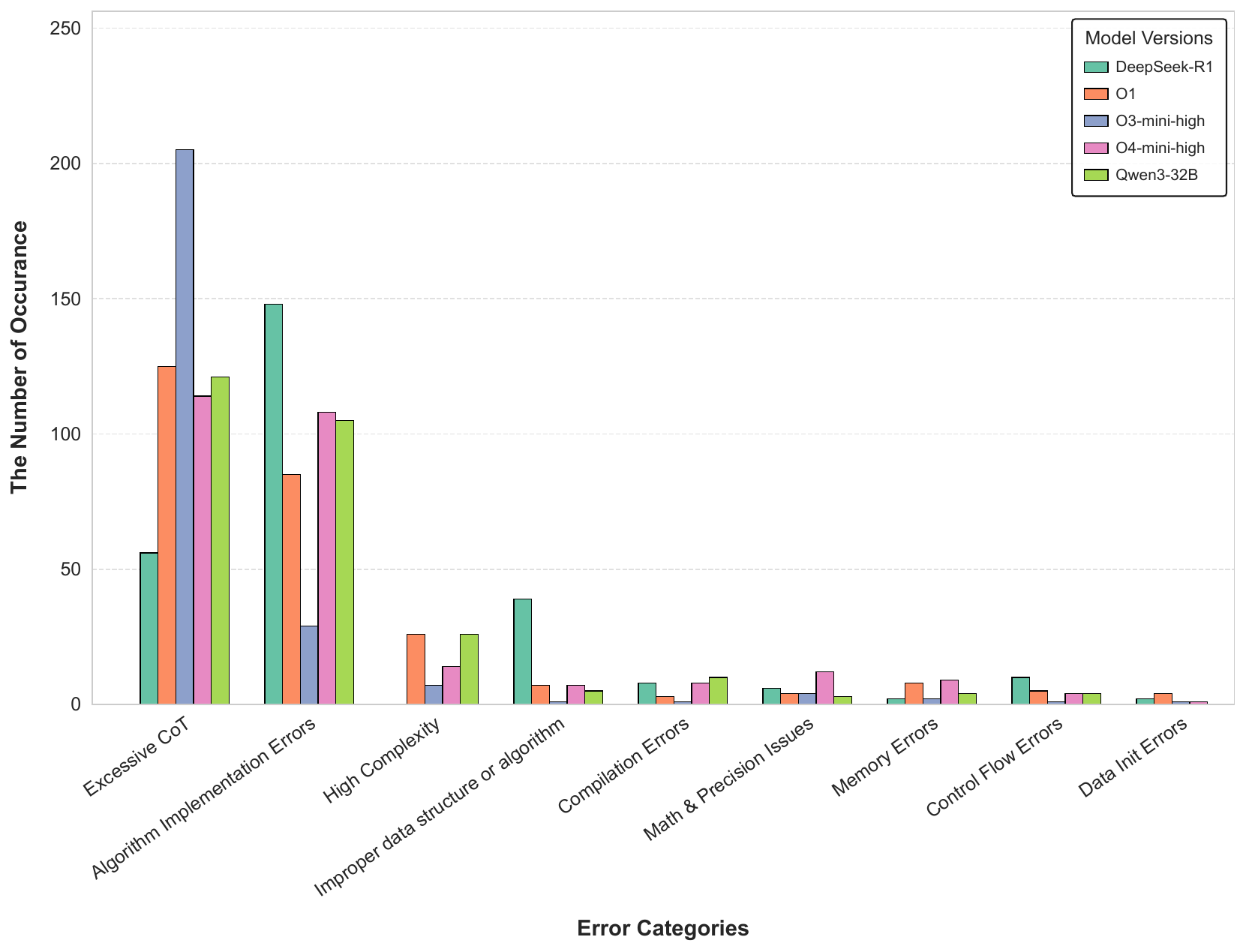}
    \caption{The error type for each model.}
    \label{figure/error_distribution_final}
\end{figure}

The error type are shown in Fig~\ref{figure/error_distribution_final}, with analysis shown in Section~\ref{sec:QA}.

\subsection{Participant Recruitment and Compensation}
\label{sec:Recruitment}
For the six Informatics Olympiad participants, they are all undergraduate students from Chinese universities with a computer science background, and have achieved at least a silver medal in the China Collegiate Programming Contest(CCPC)\footnote{\url{https://ccpc.io/}}. They work on a part-time basis.

As for the three translators, they all hold at least a bachelor's degree in Master of Translation and Interpreting for English translation and have a minimum of two years of professional experience in translation. They work full-time with a daily workload of less than 8 hours.

For crowd-sourced participants of human-comparable evaluation, we publish competitions on the Internet\footnote{\url{https://agi-eval.cn/competition/activity}} and invite CCPC student contestants from various Chinese universities. Each competition offers subsidies to participants.

All the above personnel receive hourly wages (with crowd-sourced participants' subsidies calculated based on competition time) or daily wages in accordance with local labor regulations.

\subsection{Prompts}
\label{sec:prompt}

We list the prompts used in the paper from Listing~\ref{lst:Frist} to \ref{lst:last}.

\clearpage

\begin{figure*}
\centering
\footnotesize
\begin{lstlisting}

Please use {language} language to solve the following problem. Please use the OJ's input and output format:

```
{problem description and sample input/output}
```


\end{lstlisting}
\captionof{lstlisting}{Full problem prompt of OIBench for Instruction Models in English.}\label{lst:Frist}
\end{figure*}

\begin{figure*}
\centering
\footnotesize
\begin{lstlisting}

Please use {language} language to solve the following problem. Please use the OJ's input and output format:

```
{problem description and sample input/output}
```

```{language}

\end{lstlisting}
\captionof{lstlisting}{Full problem prompt of OIBench for Base Models in English. A markdown syntax is added to aid code generation.}
\end{figure*}

\begin{figure*}
\centering
\footnotesize
\begin{lstlisting}

Please use {language} language to solve the following problem. Please use the OJ's input and output format:

```
{problem description and sample input/output}
```

Pseudo code solution for the problem.
```
{Pseudocode}
```

\end{lstlisting}
\captionof{lstlisting}{Full problem prompt of OIBench Pseudo for Instruction Models in English.}
\end{figure*}

\begin{figure*}
\centering
\footnotesize
\begin{lstlisting}

Please use {language} language to solve the following problem. Please use the OJ's input and output format:

```
{problem description and sample input/output}
```

Pseudo code solution for the problem.
```
{Pseudocode}
```

```{language}
\end{lstlisting}
\captionof{lstlisting}{Full problem prompt of OIBench Pseudo for Base Models in English.  A markdown syntax is added to aid code generation.}
\end{figure*}

\begin{figure*}
\centering
\footnotesize
\begin{lstlisting}
I'll give you a problem and its code, the code has some bugs, please fix the code according to the problem, and return the fixed code.

Task:
```md
{problem}
```


Code:
```cpp
{code}
```


\end{lstlisting}
\captionof{lstlisting}{Full problem prompt of Bugfix task.}
\end{figure*}

\begin{figure*}
\centering
\footnotesize
\begin{lstlisting}

I'll give you a problem and its code, the code has some missing parts, please complete the code according to the problem, and return the complete solution.

Task:
```md
{problem}
```

Code:
```cpp
{code}
```

\end{lstlisting}
\captionof{lstlisting}{Full problem prompt of Code Complete task.}
\end{figure*}

\begin{figure*}
\centering
\footnotesize
\begin{lstlisting}

please translate the following C++ code to {language} code

```cpp
{code}
```

\end{lstlisting}
\captionof{lstlisting}{Full problem prompt of Code Translate task.}
\end{figure*}

\begin{figure*}
\centering
\footnotesize
\begin{lstlisting}

Below is a C++ code and its corresponding input. Please guess its output based on the input. Please return the output using markdown syntax.
```cpp
{code}
```

```
{input_block}
```

\end{lstlisting}
\captionof{lstlisting}{Full problem prompt of Code Interprete task.}\label{lst:last}
\end{figure*}

\clearpage
\newpage

\bibliographystyle{unsrt}
\bibliography{references}

\clearpage
\newpage

\end{CJK}

\end{document}